\pdfoutput=1

\documentclass[11pt]{article}

\usepackage[review]{EACL2023}
\usepackage{booktabs,chemformula}
\usepackage{tabularx}
\usepackage{times}
\usepackage{latexsym}
\usepackage{epsfig,graphicx}
\usepackage{stfloats}

\usepackage[T1]{fontenc}

\usepackage[utf8]{inputenc}

\usepackage{microtype}

\usepackage{inconsolata}

\title{EMP-EVAL: A Framework for Measuring Empathy in Open Domain Dialogues}

\author{Bushra Amjad , Muhammad Zeeshan, Mirza Omer Beg \\
  Department of Artificial Intelligence and Data Science\\
  National University of Computer and Emerging Sciences \\
  Islamabad, Pakistan \\
  \texttt{\{bushra.amjad,i191711,omer.beg\}@nu.edu.pk}
  }

\begin{document}
\maketitle
\begin{abstract}
Measuring empathy in conversation can be challenging, as empathy is a complex and multifaceted psychological construct that involves both cognitive and emotional components. Human evaluations can be subjective, leading
to inconsistent results. Therefore, there is a need
for an automatic method for measuring empathy
that reduces the need for human evaluations. In this paper, we proposed a novel approach EMP-EVAL, a simple yet effective automatic empathy evaluation method. The proposed technique takes the influence of Emotion, Cognitive and Emotional empathy. To the best knowledge, our work is the first to systematically measure empathy without the human-annotated provided scores. Experimental results demonstrate that our metrics can correlate with human preference, achieving comparable results with human judgments.

\end{abstract}

\section{Introduction}

Empathy is a vital component of human communication, and it is increasingly being recognized as an important aspect of conversational agents and chatbots. Empathy refers to the ability to understand and share the feelings of others, which is crucial in building trust and rapport with users. However, measuring empathy in open-domain dialogues is a challenging task, as it requires understanding the context and emotions of the conversation. In recent years, there has been a growing interest in developing automatic methods for measuring empathy in open-domain dialogues \cite{barriere-etal-2022-wassa} \cite{hosseini-caragea-2021-distilling-knowledge}. Traditional methods for measuring empathy rely on human evaluations, which can be time-consuming and costly. Additionally, human evaluations can be subjective, leading to inconsistent results. Therefore, there is a need for an automatic method for measuring empathy that reduces the need for human evaluations.

Human-based evaluation can be challenging to set up and carry out \cite{VANDERLEE2021101151}. It is important to carefully design and instruct the experiment to ensure that it accurately reflects real-world conditions. This may involve selecting a representative sample of users, providing clear instructions on how to complete the tasks, and carefully controlling for variables that could influence the results. One way to address the high variability of user behavior is to use statistical techniques to analyze the data and account for individual differences among users. It is also important to carefully consider the sample size and ensure that it is large enough to provide reliable results  \cite{VANDERLEE2021101151}. Measuring empathy in conversation is challenging, as empathy is a complex and multifaceted psychological construct that involves both cognitive and emotional components. 

\begin{figure}[t]
\centering
\includegraphics[width=0.5\textwidth]{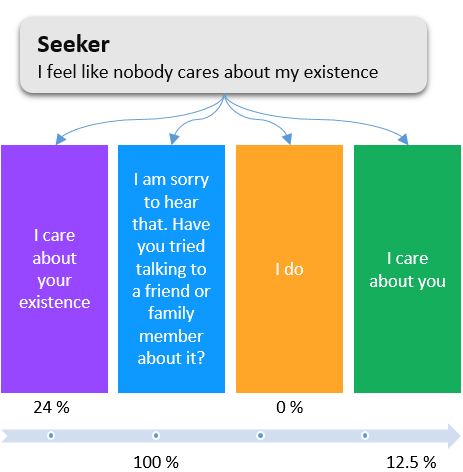}
\caption{An example of multiple response posts with the same seeker post. Each response reflects the same emotional state but a different empathy level. }
\end{figure}





Empathy is a complex and multi-faceted construct that encompasses cognitive, emotional, and behavioral components \cite{doi:10.1177/1754073914558466}. In open-domain dialogues, it is particularly challenging to measure empathy, as the conversations are not constrained to a specific topic or domain. Emotional empathy refers to the ability to experience and share the emotions of others. This type of empathy is often characterized by an automatic emotional response to the feelings of others and may involve feelings such as compassion, concern, or even distress. Cognitive empathy, on the other hand, refers to the ability to understand and comprehend the perspectives and thoughts of others \cite{smith2006cognitive}. This type of empathy involves the use of reasoning and conceptual understanding to interpret and make sense of the experiences and emotions of others. Cognitive empathy allows us to put ourselves in the shoes of others and to understand their thoughts, feelings, and motivations. Both emotional and cognitive empathy is important for effective social interactions and relationships, and they often work together to allow us to understand and connect with others. While emotional empathy is often thought of as more automatic and intuitive, cognitive empathy requires more effort and conscious thought and can be developed and improved through learning and practice.

It is important to note that no single method is likely to be sufficient on its own for measuring empathy in conversation, and a combination of approaches may be necessary to provide a more comprehensive assessment. Empathy prediction in natural language processing (NLP) refers to the use of machine learning techniques to identify and predict empathetic responses in text. This can be useful in various applications, such as customer service, where identifying and responding to customer emotions can improve the overall experience \cite{DBLP:conf/lrec/SedocBNBU20}. There have been a number of studies on empathy prediction in NLP. 

Previous studies focus on machine learning algorithms to classify text as empathetic or non-empathetic in a supervised manner. This is done using a variety of features, such as the use of certain words or phrases, the tone of the language, and the presence of emotional cues. Other studies have focused on the use of deep learning techniques, such as recurrent neural networks (RNNs) and transformer models, to predict empathy in text. These approaches have shown promising results, with some studies reporting high levels of accuracy in empathy prediction \cite{chen-etal-2022-iucl-wassa}, \cite{lahnala-etal-2022-caisa-wassa} and \cite{hosseini-caragea-2021-distilling-knowledge}.

Overall, the literature suggests that empathy prediction in NLP is a challenging but important task, with the potential to improve the effectiveness of communication in a variety of settings. However, more research is needed to understand the underlying mechanisms fully and to develop robust and reliable models for empathy prediction. Deep Neural Networks have made significant advances in natural language processing and pattern analysis in recent years. They have also been used to train corpora, which can provide a data-based dialogue system with more depth and generalization \cite{li-etal-2017-dailydialog}.
In this paper, we propose a novel approach to automatically measure Empathy in open-domain dialogues considering the high variability of user behavior. The proposed technique can be used to improve empathetic Dialogue generation. We proposed an emotion scale that is helpful in modeling empathy as per social science experts. We tested our metric on state-of-the-art dialogue generation models and provided a comparative analysis that can be helpful in future research for dialogue models. 

\section{Related work}

Evaluating empathy in dialogue is a challenging task in Natural Language Processing (NLP) due to the complexity and subjectivity of the concept. Empathy is a multidimensional construct that encompasses both cognitive and affective aspects, such as understanding others' perspectives, emotions, and feelings. One of the most commonly used evaluation techniques in NLP is a model-based approach that uses machine learning to classify dialogue as empathetic or non-empathetic. In this approach, a dataset of labeled dialogues is used to train a model. This approach has been used in several studies \cite{phdthesis}, \cite{buechel-etal-2018-modeling}, \cite{chen-etal-2022-iucl-wassa}, \cite{lahnala-etal-2022-caisa-wassa} and \cite{hosseini-caragea-2021-distilling-knowledge}. Another evaluation technique is the to detect the emotions expressed in dialogue. This approach has been used in several studies \cite{canales2014emotion} and \cite{ragheb2019attention}. Additionally, word embeddings have been used to analyze empathy in different dialogues by comparing the similarity of the words used in the dialogues. Lexical-based approaches are also followed by some researchers \cite{DBLP:conf/lrec/SedocBNBU20} to predict word-level ratings that are related to empathy. Recently, a study was conducted to understand empathy in conversation. For this purpose, \cite{sharma2020empathy} have presented a framework to predict the factors on which empathy depends and solved this as a classification problem.
 
 Despite these methods having shown promising results, it's important to note that all these methods have their own limitations. A major limitation of these approaches is that it relies on labeled data, which may not always be available or may not align with human understanding of empathy. In order to measure empathy automatically no single method is likely to be sufficient on its own for measuring empathy in conversation, and a combination of approaches is necessary to provide a more comprehensive assessment.

\section{Measuring Empathy in Dialogues}
\begin{figure*}[t]
    \centering
    \includegraphics[width=1\textwidth]{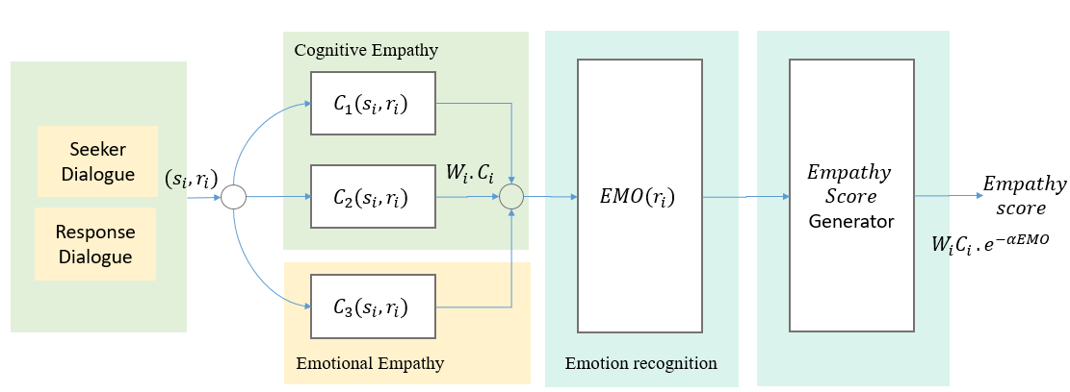}
    \caption{Our proposed automatic empathy evaluation framework for open domain dialogues. Our approach is  divided into three stages. First is empathy category identification from a seeker and response post pair. It includes the identification of cognitive and emotional empathy factors. Second is emotion identification of the dialogue and Lastly, this all is fed into the empathy score generation function.
}
    \label{fig:frame}
    \end{figure*}

We propose an automatic evaluation metric called EMP-EVAL to automatically measure empathy in dialogues. We used the EPITOME-based metric used in previous studies \cite{sharma2020empathy} as a backbone for empathy category identification. We analyze whether they capture all forms of empathy in the form of dialogue acts or not. Based on the analysis, we propose a new evaluation method called EMP-EVAL (Figure \ref{fig:frame}).

\subsection{Problem Statement}
Given a dialogue context \textit{c = {c1,c2...cn}} and a response \textit{r = {r1,r2...rn}} our goal is to learn a function \textit{f : (c, r) → s} that predicts the score for each category. Let \textit{Si = si1, ..., sim} be a seeker post and \textit{Ri =ri1, ..., rin} be a corresponding response post. For the pair \textit{(Si, Ri)}, we want to perform three tasks i.e, Empathy Category Identification, Emotion Recognition and Empathy scoring.

\subsection{Empathy Category Identification}
\noindent


\noindent Research has shown that there is a relationship between dialogue acts and empathy in that certain dialogue acts can be associated with empathetic behavior. For example, a dialogue act that involves making a statement about the other person's feelings or acknowledging the other person's feelings can be considered an empathetic act. Similarly, dialogue acts that involve making a request for help or advice, or offering help or advice, can also be associated with empathetic behavior (Table \ref{fig:emotionmapping}).

We studied the relationships between dialogue acts and empathy. We discover that the act of disgusted, disapproving, and advising dialogue does not reflect empathy behavior in dialogues as shown in Table \ref{fig:emotionmapping}.

\begin{table}[ht]
\begin{tabularx}{\linewidth}{p{1.7cm}p{3.5cm}X}
\toprule 
    & Dialogue Act & Value Range \\
   \midrule
   Category 1 & Wishing & \{0,1,2\} \\
      & Sympathizing\\ 
      & Consoling \\ 
      & Expressing care or concern & \\
      & Acknowledging & \{0,1,2\} \\
        & Appreciating \\
        & Encouraging   \\
    Category 2 & Questioning & \{0,1,2\} \\
      & Exploring &  \\
    Category 3 & Sharing own thoughts & \{0,1,2\} \\
               & Sharing own opinion \\
               & Sharing own experience \\
               & Relating to own experience  \\
   
    \hline

\end{tabularx}
\caption{Dialogue Acts with their empathetic behaviours along with their value ranges that have been used in the process.}
\label{fig:dialactcategories}
\end{table}

\begin{table*}[ht]
\begin{tabularx}{\linewidth}{p{3cm}p{7cm}X}
\toprule 
   & Dialogue Act & Behaviour  \\
   \midrule
    Category 1 & Questioning & Empathetic \\ 
    & Acknowledging & Empathetic\\
    & Wishing & Empathetic\\
    & Sympathizing & Empathetic\\
    & Agreeing & Empathetic\\
    & Appreciating & Empathetic\\
    & Consoling & Empathetic\\
    & Encouraging & Empathetic\\
    & Expressing care or concern & Empathetic\\
    & Sharing own thoughts/opinion & Empathetic\\
    & Sharing or relating to own experience & Empathetic\\
    Category 2 & Disgusted & Non-Empathetic\\
    & Disapproving & Non-Empathetic\\ 
    & Advising & Non-Empathetic\\
    
    \hline
\end{tabularx}
\caption{Dialogue acts in category 1 reflect empathy. Other Dialogue acts like Disgusted, Disapproving and Advising are said to be Non-Empathetic.}
\label{fig:emotionmapping}
\end{table*}

\noindent EPITOME \cite{sharma2020empathy} is a conceptual framework for expressing empathy in text-based, asynchronous communication contexts. However, we noticed that some dialogue acts which contribute to empathetic behavior are not captured by their model. It lacks encouraging, acknowledging, and appreciating behavior as part of emotional empathy. To capture this behavior, we extended the dataset by more dialogue pairs that show this behavior. Categories are shown in table \ref{fig:dialactcategories}. We use RoBERTa \cite{Liu2019RoBERTaAR} to encode the context c and the response r. These encodings are then used to predict the class for emotional and cognitive empathy. Each response was measured by one of the values (0, 1, or 2) predicted from the fine-tuned language model. Higher values indicate stronger empathy and a lower value indicates lower empathy.

\begin{equation} 
    Ci(emo) = RoBERTa(ci,ri)
\end{equation}
\begin{equation} 
    Ci(cogn) = RoBERTa(ci,ri)
\end{equation}

\noindent However, it's important to note that while empathetic dialogue acts can be associated with empathetic behavior, they may not necessarily reflect a person's level of empathy. Empathy is a multidimensional construct that encompasses both cognitive and affective aspects, and therefore, it's important to use multiple metrics to evaluate empathy. Next, we evaluated the dialogue based on its emotional state.

\subsection{Dialogue Emotion Scoring} 
Emotion plays a significant role in empathy prediction, as being able to identify and understand the emotions of others is a crucial aspect of empathy. Being able to empathize with others can help to regulate one's emotions, as it allows one to understand the reasons behind others' feelings. Additionally, empathy can also lead to increased feelings of compassion and altruism, which are positive emotions associated with helping others.

\begin{figure}[ht]
    \centering
    \includegraphics[width=0.5\textwidth]{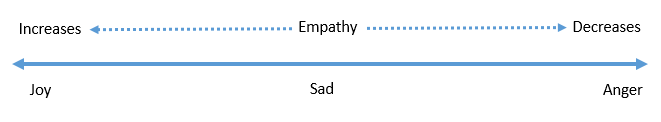}
    \caption{Our proposed Emotion Empathy scale. Emotions with a higher value penalize the empathy score. }
    \label{fig:scale}
    \end{figure}

We proposed an emotion scale (Figure \ref{fig:scale}) based on Ekman's wheel of emotion \cite{ekman1999basic}. It is a framework developed by psychologist Paul Ekman \cite{ekman1999basic} that maps out the relationships between different basic emotions. The wheel includes six basic emotions: anger, disgust, fear, happiness, sadness, and surprise, which are thought to be universal and cross-cultural. We assigned a higher value to the emotions that are non-empathetic and a lower value to the ones which show empathy behaviors (Table \ref{fig:emotionmapping}). They have an inverse relation with empathy e.g. Emotion with a higher value will penalize the empathy score. For dialogue, we predicted the emotion label and then map it to a numerical value as per the emotion scale. 

\subsection{Empathy Scoring}
\noindent To compute the empathy score, all the previously computed scores are fed into an empathy scoring function. An empathy score function is a mathematical function that assigns a numerical value to a dialogue utterance to indicate the level of empathy present in it. This score is used to evaluate the empathy level of a model. An empathy score is calculated as the weighted sum of all categories. A weight is assigned to each category depending on the impact they have on empathetic behavior (Equation \ref{eq:empathy}). A lower empathy score indicates a low empathetic dialogue and a higher empathy score indicates a more empathetic dialogue. 

\begin{equation} \label{eq:empathy}
    \sum_{i=1}^{n} Wi(Ci).\epsilon ^{-EMOi} 
\end{equation}

\begin{table*}[t]
\begin{tabularx}{\linewidth}{p{6cm}p{2cm}p{3cm}X}
\toprule 
    Name &  Year & Category & Avg Empathy Score \\
   \midrule
   Eliza \cite{weizenbaum1966eliza} & 1966 & Rule based & 3 \\
   PARRY \cite{zemvcik2019brief} & 1988 & Rule based & 3 \\
   DialoGPT \cite{https://doi.org/10.48550/arxiv.1911.00536} & 2019 & Learning based & 1.97344 \\
   FacebookBlender \cite{https://doi.org/10.48550/arxiv.1907.06616} & 2019 & Learning based & 3.5 \\
   FacebookBlender-400 \cite{https://doi.org/10.48550/arxiv.1907.06616} & 2021 & Learning based & 6.1 \\
   GPT-3 & 2022 & Learning based & 9.2 \\
   \hline

\end{tabularx}
\caption{Facebookblender from parl-AI and GPT-3 have comparatively better results. The first chatbot named ELIZA also has a good empathy score as compared to DialoGPT but it is rule-based. 
}
\label{fig:comparison}
\end{table*}

\section{Experimental Setup}

\subsection{Datasets}
We use the Mental Health Subreddits provided by \cite{sharma2020empathy} which contain high-quality open-domain conversations from Reddit threads. In addition, two datasets, DailyDialogue \cite{li-etal-2017-dailydialog} and EmpatheticDialogues \cite{rashkin-etal-2019-towards} are considered unseen datasets to verify the transferability of the metrics. 

\subsection{Dialogue Models}
We consider generation-based dialogue models to obtain responses for metric evaluation so that the performance of the metrics can be assessed comprehensively. Specifically, we first deploy facebookBlender \cite{https://doi.org/10.48550/arxiv.1907.06616} from the ParlAI platform and DialoGPT \cite{https://doi.org/10.48550/arxiv.1911.00536}which can output human-like responses.

\subsection{Implementation Details.}
For the empathy category identification task, we fine-tuned RoBERTa on our extended dataset. We used the default hyper-parameter settings provided by \cite{sharma2020empathy}. For this purpose, we fine-tuned the used RoBERTa classifiers. Their bi-encoder architecture (Humeau et al., 2019) facilitates joint modeling of (Si, Ri) pairs. Moreover, the use of attention helps in providing context from the seeker's post, Si. Following the author’s official code, we fine-tuned three RoBERTa base models to measure the empathy category. For the emotion recognition task, we used the pre-trained T5 language model \cite{https://doi.org/10.48550/arxiv.1907.06616}.

\subsection{Human Judgements}
It is common to use human annotations as a way to evaluate the performance of dialogue models, especially when the automatic evaluation metrics are not sufficient or do not accurately reflect the quality of the model's responses. We compared the results of our metric to the judgments made by humans to see if there was a correlation between the two. The individuals were given samples from each model and were asked to rate their empathy on a scale of 0 to 10. 

\section{ Experimental Results}

The Pearson correlation between the human evaluations and our predicted score is 0.72. We have seen that the Pearson correlation scores produced
\cite{barriere-etal-2022-wassa}.

by our approach are comparatively higher as compared to the previous studies done on human-annotated scores for learning. SINAI, CAISA, and LingJing are teams belonging to WASSA 2022 shared task \cite{barriere-etal-2022-wassa} (Table \ref{fig:results}). We tested six models during inference time. These scores suggest that the GPT-3 \cite{brown2020language} exhibits more empathetic behavior as compared to other models. The average of the obtained empathy scores is reported in Table \ref{fig:comparison}.

\begin{figure}[t]
    \centering
    \includegraphics[width=0.5\textwidth]{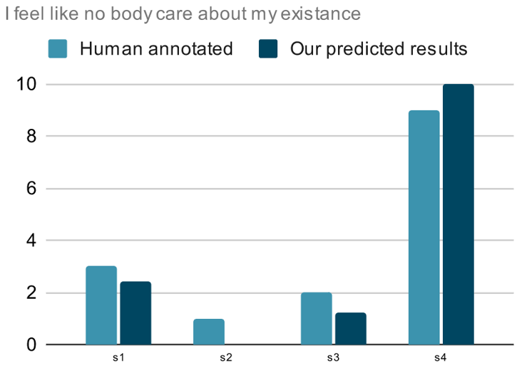}
    \caption{Comparison of Human annotations and our predicted empathy scores}
    \label{fig:ex1}
\end{figure}

 \begin{figure}[t]
    \centering
    \includegraphics[width=0.5\textwidth]{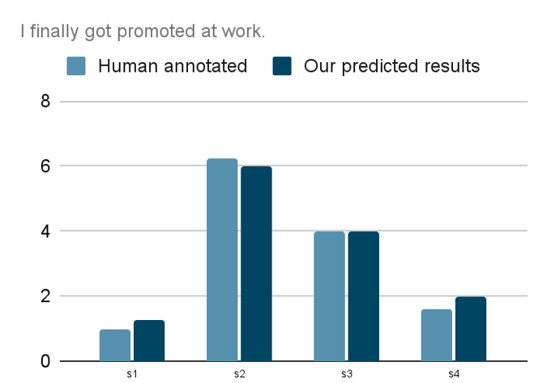}
    \caption{Comparison of Human annotations and our predicted empathy scores}
    \label{fig:ex2}
 \end{figure}

\noindent Examples are shown in Figure \ref{fig:ex1} and \ref{fig:ex2} for comparison between our predicted scores and human annotated values. 
Figure \ref{fig:ex1} illustrates the following example. Seeker post: "I feel like nobody cares about my existence." Response Post 1: "I care about your existence." Response post 2: "I do." Response Post 3: "I care about you." Response Post 4: "I’m sorry to hear that." Have you tried talking to a friend or family member about your feelings?" Figure \ref{fig:ex2} illustrates the following example. Seeker post: "I finally got promoted at work." Response Post 1: "Congrats!"
Response Post 1: "Congratulations! I hope you enjoy your new position. What do you do for
work?"  Response Post 1: "Congratulations! What do you do for a living?" Response Post 1: "Congrats on the promotion!" In the bar graph (Figure \ref{fig:ex1} and \ref{fig:ex2}) light blue color represents human
annotated scores and dark blue represents our automatically predicted results. The correlation between our prediction and ground truth values is 0.72 which shows a moderate linear relationship between the two approaches.

\begin{table}[t]
\begin{tabularx}{\linewidth}{p{0.7cm}XXXX}
\toprule 
    & SINAI & CAISA & LingJing & \textbf{EMP-EVAL} \\
   \midrule
   Score& 0.541 & 0.524 & 0.508 & \textbf{0.72} \\
   \hline

\end{tabularx}
\caption{Pearson Correlation of Predicted Empathy scores}
\label{fig:results}
\end{table}

\section{Conclusion and Future Work}
Measuring empathy in conversation can be challenging, as empathy is a complex and multifaceted psychological construct that involves both cognitive and emotional components. We proposed a novel approach to automatically measure Empathy in open-domain dialogues considering the high variability of user behavior. The proposed technique can be used to improve the empathetic Dialogue generation process. We proposed an emotion scale that is helpful in modeling empathy as per social science experts. We tested our metric on state-of-the-art dialogue generation models and provided a comparative analysis that can be helpful in future research for dialogue models. 

Additionally, more sophisticated machine learning algorithms can be used to improve the accuracy and robustness of the proposed method. Furthermore, the proposed method can be integrated into conversational agents and chatbots.

\bibliography{custom}
\bibliographystyle{acl_natbib}
\end{document}